\documentclass{svproc}

\usepackage{url}
\usepackage{cite}
\usepackage{enumitem}
\usepackage{amsmath,amssymb,amsfonts}
\usepackage{algorithmic}
\usepackage{graphicx}
\usepackage{textcomp}
\usepackage{xcolor}
\usepackage{booktabs}  %
\usepackage{multirow}  %
\usepackage{siunitx}

\usepackage{caption}   %
\usepackage{hyperref}  %

\graphicspath{{Images/}}

\begin{document}

\mainmatter              %
\title{Robotic Perception with a Large Tactile-Vision-Language Model for Physical Property Inference}

\titlerunning{Cross-Modal Robotic Perception for Physical Property Inference}  %

\author{
	Zexiang Guo\inst{1}\textsuperscript{*} \and 
	Hengxiang Chen\inst{1}\textsuperscript{*} \and 
	Xinheng Mai\inst{1}\textsuperscript{*} \and 
	Qiusang Qiu\inst{1} \and 
	Gan Ma\inst{2} \and
        Zhanat Kappassov \inst{3} \and
	Qiang Li\inst{1}\textsuperscript{\dag} \and 
	Nutan Chen\inst{4}
}

\authorrunning{ Zexiang Guo et al.} %
\tocauthor{Zexiang Guo, Hengxiang Chen, Xinheng Mai, Qiusang Qiu,
	Gan Ma,Zhanat Kappassov, Qiang Li, and Nutan Chen}

\institute{College of Big Data and Internet, Shenzhen Technology University, China,\\
	\and Sino-German College of Intelligent Manufacturing, Shenzhen Technology University, China,\\
	\and Robotics Department, Institute of Smart Systems and Artificial Intelligence (ISSAI), Nazarbayev University, Kazakhstan\\
	\and Foundation Robotics Labs, Germany}

\maketitle              %
\textsuperscript{*}These authors contributed equally to this work.
\textsuperscript{\dag}Corresponding author

\renewcommand{\thefootnote}{}
\footnotetext{International Conference on Climbing and Walking Robots (CLAWAR), 2025}
\renewcommand{\thefootnote}{\arabic{footnote}}

\begin{abstract}
	
	Inferring physical properties can significantly enhance robotic manipulation by enabling robots to handle objects safely and efficiently through adaptive grasping strategies. Previous approaches have typically relied on either tactile or visual data, limiting their ability to fully capture properties. 
	We introduce a novel cross-modal perception framework that integrates visual observations with tactile representations within a multimodal vision-language model. Our physical reasoning framework, which employs a hierarchical feature alignment mechanism and a refined prompting strategy, enables our model to make property-specific predictions that strongly correlate with ground-truth measurements.  
	Evaluated on 35 diverse objects, our approach outperforms existing baselines and demonstrates strong zero-shot generalization.
\end{abstract}
\begin{figure*}[htbp]
	\centering
	\includegraphics[width=0.8\textwidth]{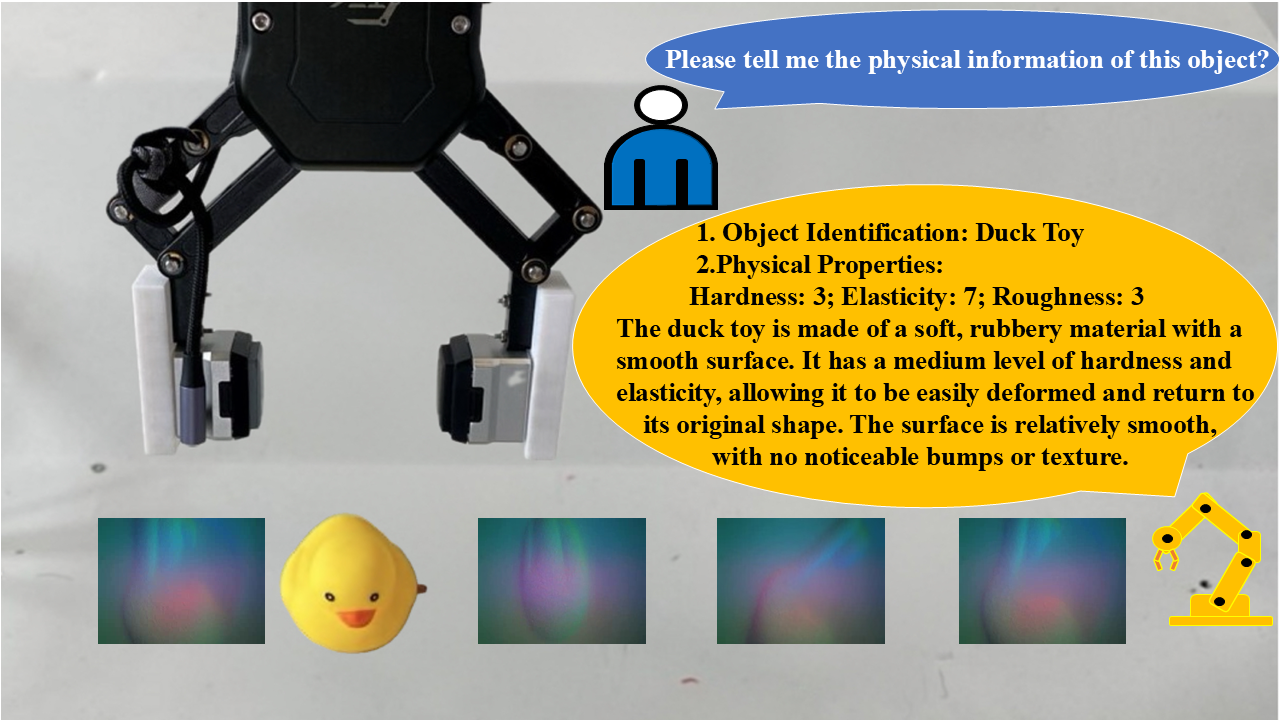} 
	\caption{Through visual and tactile image input and human language interaction, our model infers and gives detailed physical properties of the duck toy and gives specific physical property scores as specified by the structured scoring guidelines.}
	\label{fig:fig1}
\end{figure*}
\vspace{-2em}  %
\section{Introduction}
Accurate perception of object physical properties is fundamental for robots to perform reliable manipulation in unstructured environments. While humans seamlessly integrate visual and tactile cues to infer material characteristics \cite{jeka2000}, robotic systems often struggle to achieve comparable performance due to limitations in unimodal sensing. Traditional vision-based approaches, though effective for geometric perception, frequently fail to capture intrinsic material attributes such as hardness, elasticity, and surface roughness \cite{fleming2014}. Conversely, tactile sensing provides rich contact information but requires physical contact — a significant drawback when handling delicate or unknown objects \cite{chi2018}.
Recent advances in multimodal learning have shown significant potential for integrating tactile perception with language models to enhance physical reasoning capabilities \cite{zeng2023}. However, two fundamental limitations persist: (1) \textbf{Sensory constraints in tactile systems:}  Current tactile sensors offer insufficient data capture for comprehensive material characterization, particularly when handling objects with complex composite structures; and (2) \textbf{Underutilized language model potential:} Existing implementations fail to fully leverage the reasoning capacity of language models through strategic prompting and effective multimodal fusion.
To address these challenges, we propose an enhanced multimodal framework with two core innovations that enable physical property inference for robotic grasping tasks. As illustrated in Fig.\ref{fig:fig1}, our vision-tactile-language integration empowers the robotic arm to accurately estimate critical material characteristics (e.g., hardness, elasticity, surface roughness), allowing it to grasp the duck toy while preserving its structural integrity. The framework's technical advancements include:

\begin{itemize}
	\item \textbf{Proactive Perception Architecture:} By fusing visual cues with historical tactile information, our model is capable of predicting important physical attributes---such as hardness, elasticity, and roughness---prior to contact.
	\item \textbf{Structured Reasoning Prompts:} A staged reasoning protocol that guides multimodal language models through object recognition, material analysis, and property quantification to enhance inference accuracy.
\end{itemize}

\section{Related Work}
\subsection{Tactile Perception in Robotics}
Tactile sensing has become essential for robotic manipulation, with various sensor technologies capturing detailed contact information~\cite{liu2022,vanhoof2016}. Vision-based tactile sensors (e.g., GelSight \cite{yuan2017}, GelSlim\cite{donlon2018}) excel at local texture and hardness estimation but are inherently limited in capturing global object properties due to restricted sensing areas. Recent tactile representation learning approaches leveraging deep networks, including tactile-kinematic fusion for shape reconstruction~\cite{smith2020} and self-supervised tactile-visual alignment~\cite{dave2024}, have improved property estimation. %
Nevertheless, each modality individually faces limitations: tactile sensors suffer from partial observability and difficulties in dynamic property inference, whereas vision alone lacks fine-grained contact information. Previous works have addressed some of these issues by combining visual and tactile sensing~\cite{xu2023_11,taunyazov2020,li2023}. However, our approach further enhances this multimodal integration, effectively leveraging visual priors to enrich tactile perception.

\subsection{Multimodal Fusion Approaches}
The integration of tactile and visual modalities has evolved through several fusion paradigms to address limitations inherent in single-modal perception. Early fusion methods \cite{li2024}, which directly concatenate raw tactile and visual features, face performance degradation due to modality misalignment. Subsequently developed late fusion techniques \cite{liu2016} process each modality independently, yet they are limited in capturing essential cross-modal correlations required for inferring complex physical properties. More advanced hybrid methods adopt intermediate fusion strategies, including contrastive learning for feature alignment \cite{meyer2020} and attention mechanisms for adaptive modality weighting \cite{wong2023}. However, these techniques typically rely on extensive paired training datasets, potentially limiting their generalization capabilities when encountering novel objects or properties.Building upon these limitations, our research introduces a novel hierarchical prompting strategy utilizing pre-trained vision-language models as robust knowledge priors. This framework implements property-specific fusion rules, effectively enabling zero-shot generalization through structured physical reasoning.

\subsection{Physical Property Reasoning with Large Models}
Recent advances in large vision language models have demonstrated their capability for physical property reasoning by leveraging multimodal inputs and commonsense knowledge \cite{gao2024}. For instance, GPT-4V has been shown to infer liquid viscosity by analyzing time-series plots of force/torque sensor data \cite{lai2024}, while OCTOPI \cite{yu2024} predicts material properties like hardness and roughness from tactile images through specialized prompting strategies. However, existing methodologies predominantly focus on passive interpretation of sensory data without actively guiding the reasoning processes and rely on generalized multimodal fusion rather than explicitly structured, property-centric prompting. Our research substantially advances this direction by proposing a hierarchical reasoning framework that systematically decomposes physical property inference into sequential, interpretable stages-object recognition, material analysis, and quantitative assessment. Through carefully designed property-specific prompts, our approach actively directs model attention toward relevant sensory cues critical to each property, such as pressure response for hardness, deformation patterns for elasticity, and surface textures for roughness evaluation, thereby improving reasoning accuracy and interpretability. 

\section{Methodology}

\begin{figure*}[htbp]
	\centering
	\includegraphics[width=0.8\textwidth]{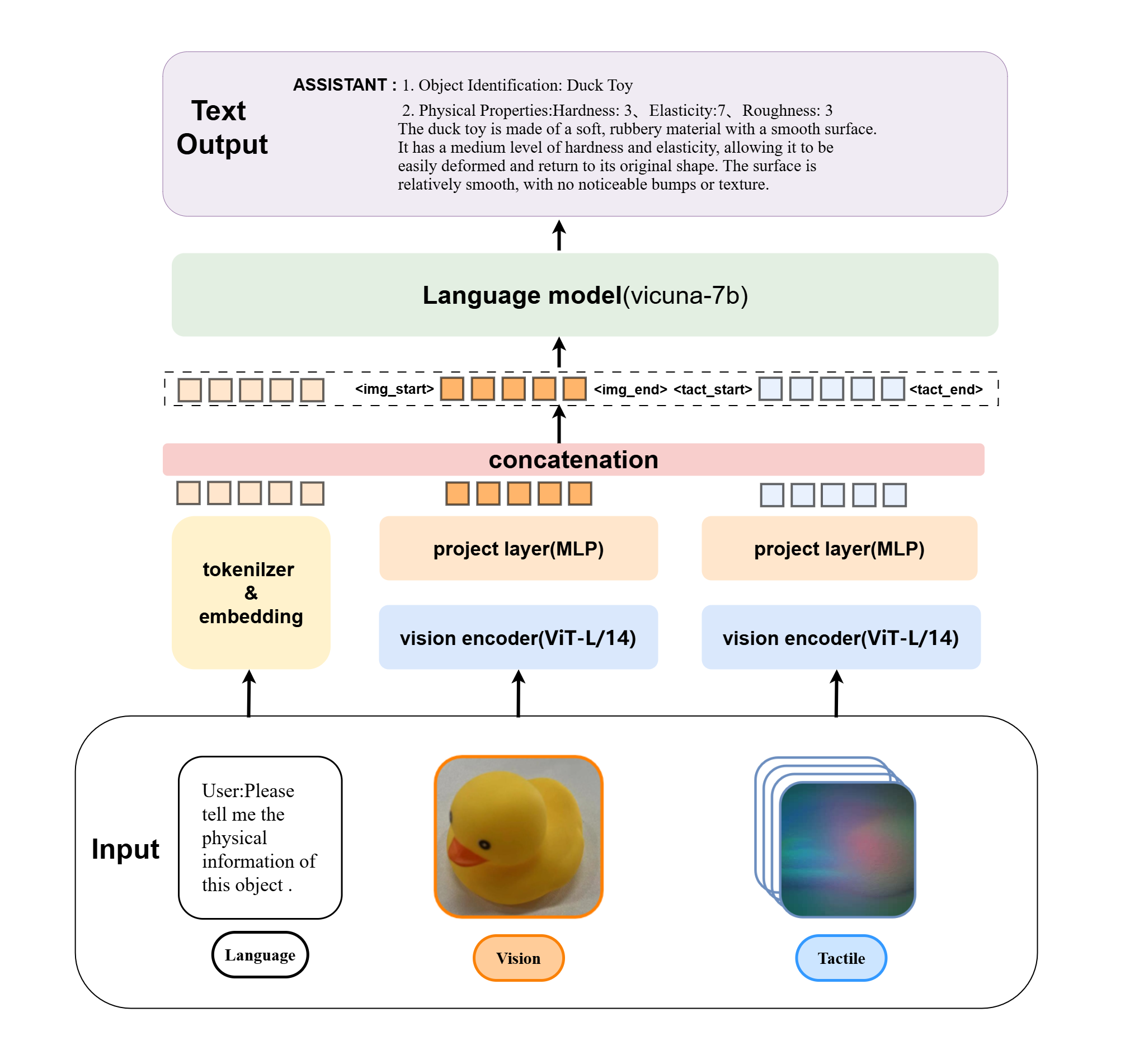} 
	\caption{The architecture of a multimodal large model. After embedding and tokenizing the object image and tactile image alongside the text, the resulting vectors are concatenated and input into the large language model.}
	\label{fig:fig2}
\end{figure*}

In our method, we introduce a multimodal model integrating textual, visual, and tactile data for comprehensive object analysis. As depicted in Fig.~\ref{fig:fig2}, the input query is parsed into dedicated modality-specific pathways. Text is tokenized and embedded via a language tokenizer, while visual and tactile images are encoded using ViT-L/14~\cite{radford2021} and projected into a shared embedding space using modality-specific MLP layers. Special markers (\texttt{<img\_start>}, \texttt{<img\_end>}, \texttt{<tact\_start>}, \texttt{<tact\_end>}) clearly delineate embedding boundaries. These embeddings are concatenated with textual features and fed into a large language model (Vicuna-7B~\cite{chiang2023}), allowing joint multimodal attention to generate detailed object property descriptions, such as hardness, elasticity, and roughness.
\subsection{Vision Processing}
\begin{figure*}[htbp]
	\centering
	\includegraphics[height=0.3\textheight]{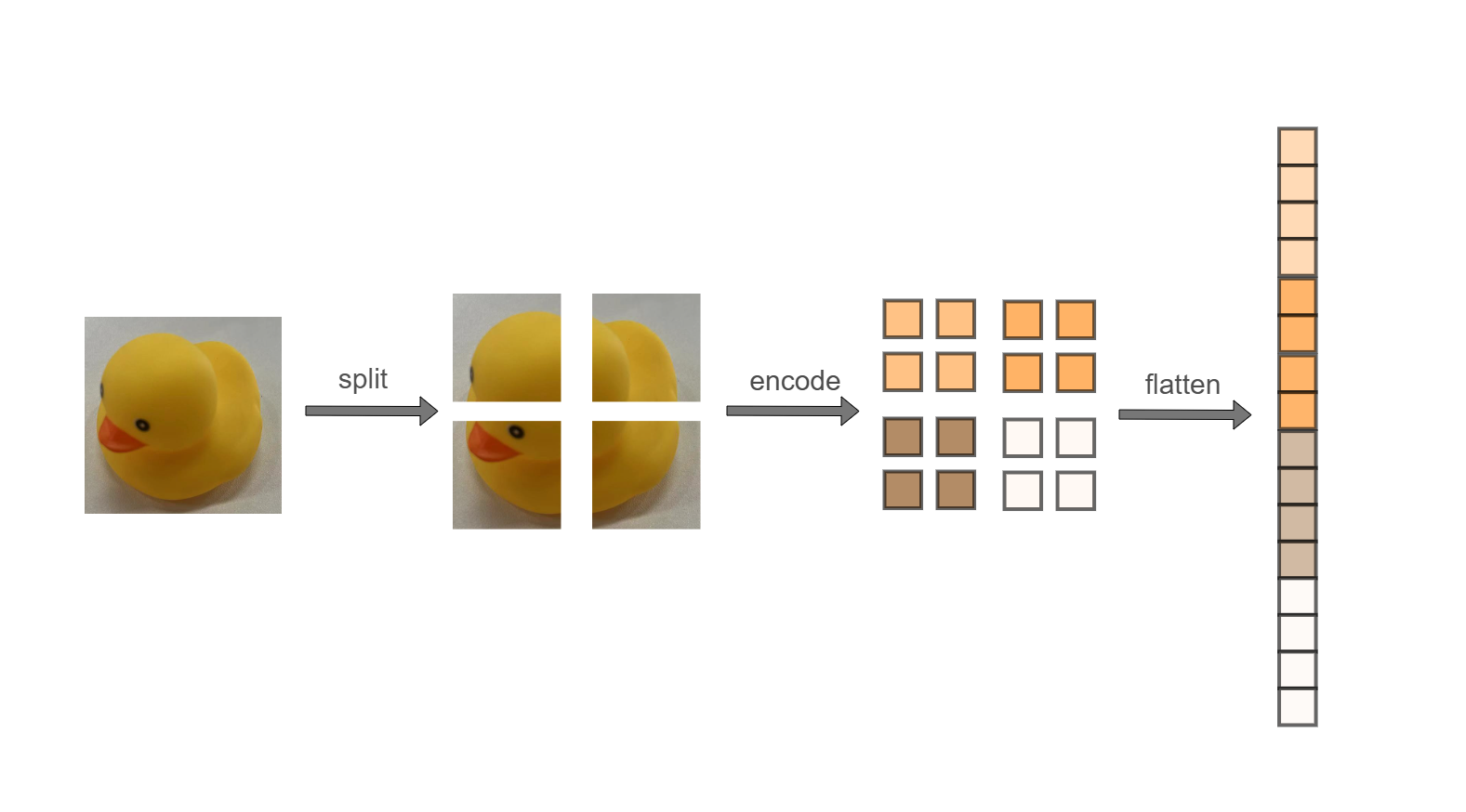} 
	\caption{Vision processing pipeline. The image is partitioned into multiple regions by the segmentation module, after which the encoder extracts a feature matrix that is flattened into a one-dimensional vector and fed into the LLM(see Fig.\ref{fig:fig2}).}
	\label{fig:fig3}
\end{figure*}\par
We employ CLIP \cite{radford2021} to process visual information, leveraging a visual encoder (ViT-L/14) trained to learn shared representations between images and text. As shown in Fig.\ref{fig:fig2} , the overall multimodal architecture incorporates image embeddings alongside textual inputs into the LLM. To align dimensionality and semantics with the LLM's native embedding space, we adopt the pre-trained linear transformation layer from LLaVA \cite{liu2023}, which projects the penultimate output of CLIP into the language model's word embedding space.\par
In addition to extracting features from the CLIP visual encoder, we insert two specialized boundary tokens, \texttt{<img\_start>} and \texttt{<img\_end>}, around image-derived embeddings. These tokens (initialized by semantic averaging of descriptive phrases and then frozen during fine-tuning) explicitly separate visual content from text inputs, thus assisting the model in distinguishing modalities. As depicted in Fig.\ref{fig:fig3}, we further segment the image into multiple regions, extract a feature matrix from each region, and flatten it into a one-dimensional representation to feed into the LLM.

\subsection{Tactile Processing}
\begin{figure*}[htbp]
	\centering
	\includegraphics[height=0.3\textheight]{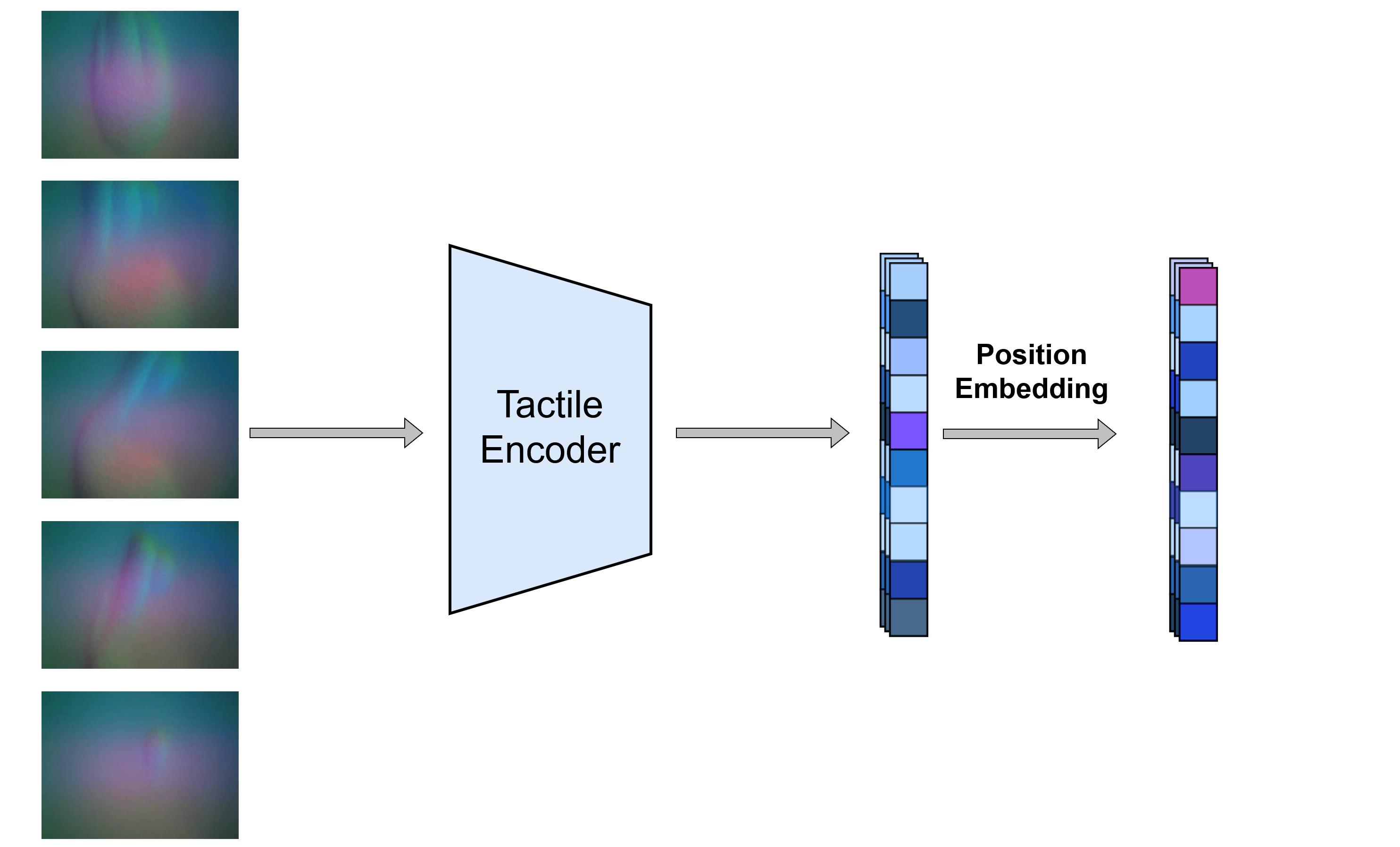} 
	\caption{A sequence of tactile images is first processed by the tactile encoder to extract feature representations. The extracted features are then transformed into a structured feature vector, followed by the addition of positional embeddings to encode temporal dependencies.}
	\label{fig:fig4}
\end{figure*}

To incorporate tactile perception and enhance physical reasoning, we adopt the OCTOPI framework. This framework employs a CLIP-based \cite{radford2021} tactile encoder that processes tactile data and fuses it with the LLM, enabling a deeper understanding of object properties. Specifically, the tactile encoder extracts features from a sequence of tactile images, encoding both spatial and temporal information. As shown in Fig.\ref{fig:fig4}, we add positional encodings to these sequential features to preserve the order and timing of tactile interactions. By training on physics-based datasets with annotated tactile videos and physical property labels, the model acquires rich, tactile-aware representations that improve performance in tasks such as object property prediction and scenario reasoning.

\subsection{Multimodal Fusion through Feature Concatenation}
After we obtain the projected object image feature vector \((F_o)\), the projected 
tactile image feature vector \((F_t)\), and the linguistic feature vector 
\((F_l)\) from the LLM's embedding space, we concatenate them channel-wise into a 
unified representation:
\[
F_{\text{concat}} = [\,F_o \; ; \; F_t \; ; \; F_l\,].
\]
This fused vector \(F_{\text{concat}}\) retains distinguishing features from each 
modality while enabling cross-modal interaction. It then serves as the input to 
downstream modules for tasks such as multimodal reasoning, classification, or object 
recognition, thereby capturing both the physical and semantic attributes of the 
target object.

\subsection{Refined Prompting Strategy for Physical Property Scoring}
\begin{table}[t]
	\centering
	\caption{Physical Property Rating Scales}
	\label{tab:rating_scales}
	\begin{tabular}{ccll}
		\hline
		\textbf{Property} & \textbf{Score Range} & \textbf{Characterization} & \textbf{Example Materials} \\
		\hline
		\multirow{5}{*}{Hardness} 
		& 1--2 & Extremely soft & Cotton, sponge \\
		& 3--4 & Soft & Rubber ball, soft plastic toy \\
		& 5--6 & Medium & Plastic container, shoe sole \\
		& 7--8 & Hard & Wood, ceramic plate \\
		& 9--10 & Extremely hard & Metal, diamond \\
		\hline
		\multirow{5}{*}{Elasticity} 
		& 1--2 & Minimal elasticity & Clay, dry sponge, wooden ruler \\
		& 3--4 & Low elasticity & Rubber eraser, hard plastic, book cover \\
		& 5--6 & Medium elasticity & Foam ball, silicone, thick rubber mat \\
		& 7--8 & High elasticity & Rubber band, bouncy ball, yoga mat \\
		& 9--10 & Maximum elasticity & Trampoline surface, latex sheet, inflated balloon \\
		\hline
		\multirow{5}{*}{Roughness} 
		& 1--2 & Extremely smooth & Glass, polished marble \\
		& 3--4 & Smooth & Plastic surface, ceramic mug \\
		& 5--6 & Medium texture & Paper, leather, cardboard \\
		& 7--8 & Rough & Sandpaper, concrete, bark of a tree \\
		& 9--10 & Extremely rough & Gravel, coarse fabric, pumice stone \\
		\hline
	\end{tabular}
\end{table}
We designed a structured prompt to enable comprehensive physical property analysis using multimodal (visual and tactile) data. The prompt clearly defines the analysis goal, emphasizing material-aware reasoning and avoiding generic responses. It guides the model through two phases: visual-based object identification (color, shape, texture) and combined material-tactile property evaluation. A 10-point Likert scale quantifies three essential properties (Table \ref{tab:rating_scales}), enhancing nuanced differentiation. Outputs include justified object identification and property scores with material rationales. Constraints ensure balanced score usage and material-focused reasoning.

When the user inputs a request along with an image, the request is encoded by the text encoder together with the prompt. Simultaneously, the object image and the tactile image are processed by their respective encoders. To facilitate proper identification and integration of different modalities, special tokens \texttt{<img\_start>}, \texttt{<img\_end>}, \texttt{<tact\_start>}, and \texttt{<tact\_end>} are used to mark the boundaries of visual and tactile features. This design enhances multimodal integration by associating visual and tactile information to provide a more comprehensive object representation. By unifying feature alignment, it improves cross-modal compatibility, while deep semantic understanding optimizes adaptation to complex scenarios.

\section{Experiments}
\subsection{Hardware}
\begin{figure*}[htbp]
	\centering
	\includegraphics[height=0.4\textheight]{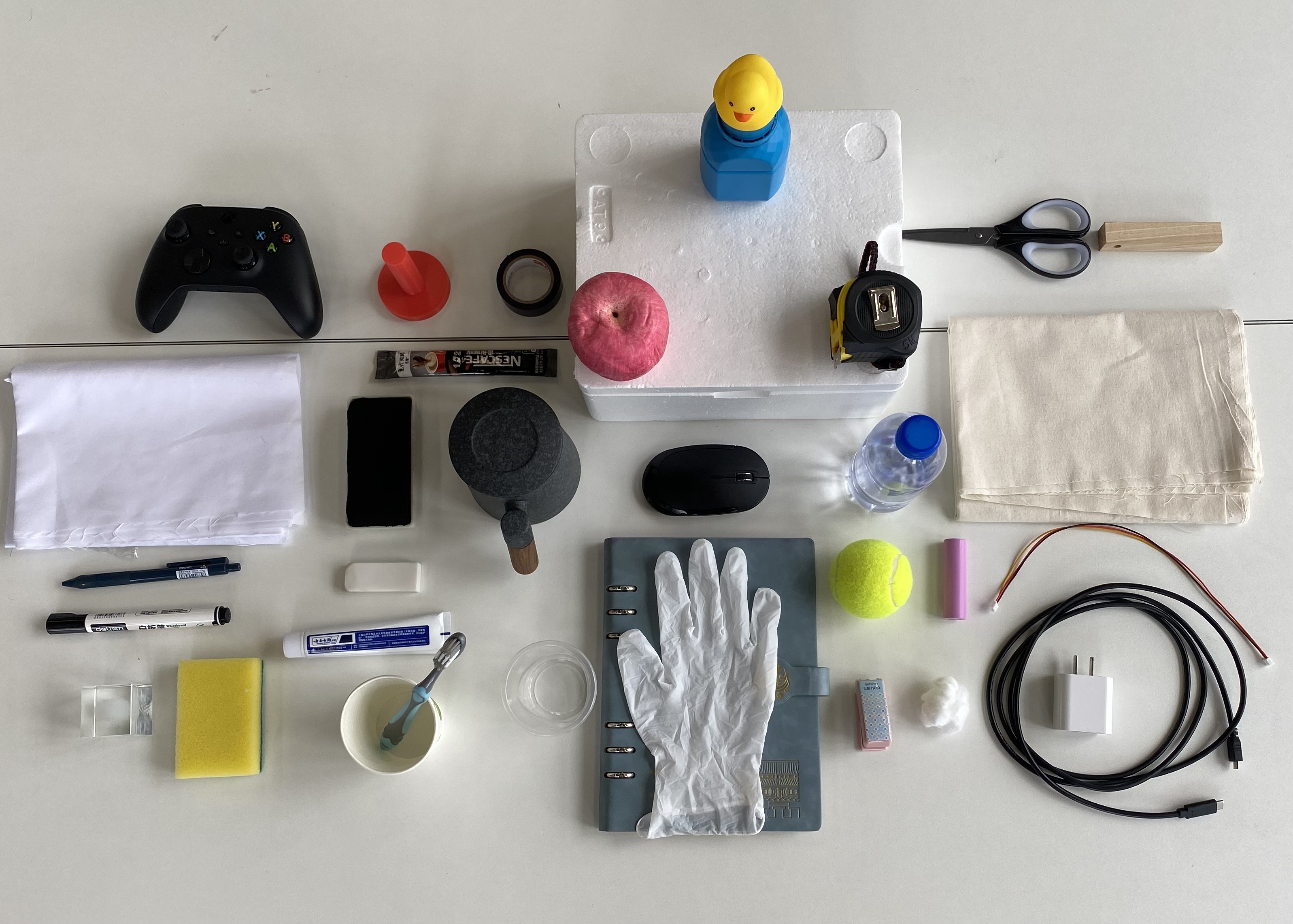} 
	\caption{Object set of 35 common household items, spanning nine major material categories—plastic, rubber, metal, wood, ceramic, glass, foam, paper, and textile—to validate the generalizability of the experimental data across diverse materials, for evaluating multimodal models in physical property reasoning.}
	\label{fig:fig5}
\end{figure*}
\begin{figure*}[htbp]
	\centering
	\includegraphics[height=0.3\textheight]{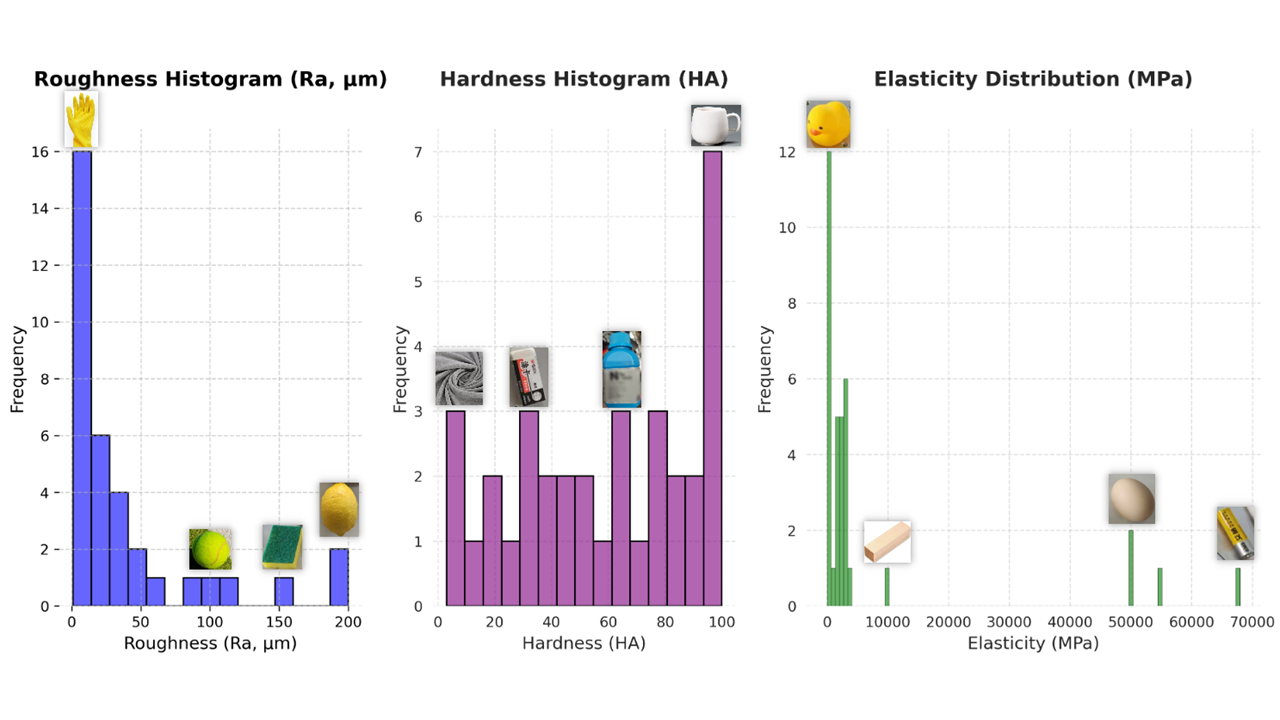} 
	\caption{During the experiment, we selected 35 objects in the laboratory, measured their roughness, hardness and elastic modulus, and drew the histogram shown in the figure above.}
	\label{fig:fig6}
\end{figure*}
To evaluate our cross-modal perception framework, we conducted comprehensive experiments using a robotic system equipped with a GelSight Mini tactile sensor for high-resolution contact data acquisition and a RealSense D410 camera for visual perception. We selected 35 common household objects (Fig.\ref{fig:fig5}) spanning diverse materials (plastic, metal, wood, rubber, etc.) and geometric properties. Each object was annotated with ground truth physical properties measured by professional instruments: hardness (Shore scale) with PosiTector SHD, elastic modulus with C610H Auto Tensile Tester, and surface roughness (Ra) with RUGOSURF 20 roughness tester (Fig.\ref{fig:fig6}).

\subsection{Data Collection}
The data collection process was designed to systematically capture multimodal information for comprehensive physical property analysis. For tactile data acquisition, we employed a GelSight Mini sensor operating at 20 fps to record six-second videos of each interaction, encompassing the complete contact cycle from approach to retraction. These videos were subsequently sampled at 250ms intervals to obtain representative frames while maintaining temporal coherence.
Ground truth measurements were obtained following established protocols to ensure accuracy and repeatability. Hardness measurements were conducted using a PosiTector SHD durometer, with three tests performed at predetermined locations on each object and a standardized 5-second dwell time. Elastic modulus characterization was performed using a C610H tensile tester, analyzing stress-strain responses in the linear deformation regime. Surface roughness was quantified with a RUGOSURF 20 profilometer, executing multiple scans per object with carefully controlled parameters.
\subsection{Experimental results}
To evaluate our model, we employed designed prompts to assess the physical properties of 35 objects through both our method and the Octopi framework\cite{yu2024}, while simultaneously obtaining ground truth measurements of the objects' attributes using specialized instrumentation. Specifically, true hardness was measured using a Shore hardness tester, the elastic modulus was determined using a universal material testing machine, and surface roughness was quantified using a portable surface roughness instrument. These instruments ensured that our ground truth data were both accurate and reproducible.

Following data collection, we normalized each dataset and computed Spearman correlation coefficients between the model scores and the ground truth measurements. These analyses allowed us to quantitatively assess the predictive accuracy of the physical property inference for hardness, elasticity, and roughness, highlighting significant differences between our approach and the baseline Octopi method.
\begin{table}[h]
	\centering
	\caption{ Zero-shot evaluation: Comparison of Spearman’s rank correlation between models and ground truth (Octopi as the tactile-only model; Octopi-ViTaL is our model)}
	\label{tab:correlation}
	\begin{tabular}{llcc}
		\toprule
		Attribute & Method & {Correlation Coefficient} & {P-value} \\
		\midrule
		\multirow{4}{*}{Hardness} 
		& Octopi-ViTaL &\textbf{0.501} &\textbf{0.005} \\
		& Octopi-ViTaL (vision only) & 0.307 & 0.099 \\
		& Octopi (fine-grained) & 0.307 & 0.099 \\
		& Octopi (original) & 0.015 & 0.935 \\
		\midrule
		\multirow{4}{*}{Elasticity} 
		& Octopi-ViTaL & \textbf{0.530} & \textbf{0.003} \\
		& Octopi-ViTaL (vision only) & 0.452 & 0.012 \\
		& Octopi (fine-grained) & 0.053 & 0.781 \\
		& Octopi (original) & -0.060 & 0.753 \\
		\midrule
		\multirow{4}{*}{Roughness} 
		& Octopi-ViTaL & \textbf{0.643} & \textbf{0.0001} \\
		& Octopi-ViTaL (vision only) & 0.413 & 0.023 \\
		& Octopi (fine-grained) & -0.010 & 0.959 \\
		& Octopi (original) & 0.118 & 0.534 \\
		\bottomrule
	\end{tabular}
\end{table}
As can be seen from the Table \ref{tab:correlation}, the correlation coefficients between the models and ground truth measurements reveal significant differences in the performance of our model compared to Octopi across the three physical attributes: hardness, elasticity, and roughness. 

For hardness, our model exhibits a moderate and statistically significant positive correlation with the ground truth (Spearman's $\rho$ = 0.501, p = 0.005), demonstrating its capability to integrate visual and tactile cues effectively. In comparison, the pure vision model yields a weaker correlation ($\rho$ = 0.307, p = 0.099), failing to reach statistical significance. Interestingly, both Octopi (fine-grained) and Octopi (original) perform worse: while the fine-grained version—Octopi  uses our prompt to score the physical properties of objects.—achieves a similar correlation to pure vision ($\rho$ = 0.307, p = 0.099), the original version—which relies on Octopi's predefined three-level classification system—shows virtually no correlation with the ground truth ($\rho$ = 0.015, p = 0.935). These results confirm that our multimodal model surpasses both unimodal and tactile-only baselines, and benefits significantly from combining sensory modalities.

In the case of elasticity, because elastic modulus is inversely proportional to perceived “elasticity,” we report the absolute value of Spearman’s $\rho$ to reflect predictive strength regardless of sign. Our model achieves a Spearman correlation of 0.530 (p = 0.003), clearly outperforming the vision-only baseline ($\rho$ = 0.452, p = 0.012). Meanwhile, Octopi (fine-grained) shows a negligible correlation ($\rho$ = 0.053, p = 0.781), and Octopi (original) displays a weak negative trend ($\rho$ = -0.060, p = 0.753). The poor performance of both Octopi variants indicates that tactile input alone lacks sufficient expressiveness for elasticity estimation, even when adapted to more descriptive prompts.             

The comparison is most striking for roughness, where our model achieves a strong and statistically robust correlation with ground truth ($\rho$ = 0.643, p = 0.0001). Although the pure vision model also yields a moderate correlation ($\rho$ = 0.413, p = 0.023), it falls short of our model's performance. Octopi (fine-grained) shows no meaningful correlation ($\rho$ = -0.010, p = 0.959), and the original Octopi version fares only slightly better ($\rho$ = 0.118, p = 0.534), with neither result statistically significant. This further demonstrates that a tactile-only approach—even with fine-tuned prompts or structured rating schemes—fails to adequately capture surface texture without visual context.

When applied in a zero-shot fashion to our new setup, the pretrained Octopi model failed to produce meaningful predictions (e.g., Spearman’s $\rho$ $<$ 0.1; see Table \ref{tab:correlation}). This failure arises from multiple domain shifts: we use a GelSight Mini with different resolution and calibration compared to Octopi’s original high-resolution GelSight; lighting and camera angles differ. These combined shifts in sensor modality, resolution, and lighting prevent Octopi from succeeding zero-shot on our data.

Overall, these results highlight the clear advantage of our multimodal approach. By fusing vision and touch, our model consistently achieves statistically significant and higher correlations with ground truth across all three physical attributes. In contrast, both the vision-only and tactile-only methods—particularly the Octopi framework in its original and adapted forms—fall short, reinforcing the value of cross-modal integration in physical property understanding.

\section{Conclusion}
We proposed a novel approach to enhance tactile perception through visual compensation and optimized prompt engineering, leveraging VLM for cross-modal robotic perception. By effectively integrating visual priors and structuring language model interactions, our method overcomes tactile-only limitations and significantly improves physical property inference, especially in roughness estimation. The success of our framework underscores the value of multimodal reasoning with VLMs for robotic applications. Future work will explore applying this multimodal tactile-visual approach to robotic grasping tasks involving adaptive manipulation of objects with different material properties.

\section*{Acknowledgment}
This research is supported by the "Natural Science Foundation of Top Talent of SZTU", 
grant no.\ GDRC202411.

\end{document}